\pdfoutput=1
\documentclass[final]{cvpr}

\usepackage{times}
\usepackage{epsfig}
\usepackage{graphicx}
\usepackage{amsmath}
\usepackage{amssymb}
\usepackage{multirow}

\usepackage[pagebackref=true,breaklinks=true,colorlinks,bookmarks=false]{hyperref}

\newlength\savewidth\newcommand\shline{\noalign{\global\savewidth\arrayrulewidth
  \global\arrayrulewidth 1pt}\hline\noalign{\global\arrayrulewidth\savewidth}}
  
\def\cvprPaperID{21} 
\def\confYear{CVPR Workshop 2021}

\begin{document}

\title{
Less is More: 
Sparse Sampling for Dense Reaction Predictions}


\author{
$\text{Kezhou Lin}^{1}$\\

\and
$\text{Xiaohan Wang}^{1}$\\

\and
$\text{Zhedong Zheng}^{2}$\\

\and
$\text{Linchao Zhu}^{2}$\\

\and
$\text{Yi Yang}^{1}$\\

\and
${}^{1}\text{Zhejiang University}\hspace{6pt} {}^{2}\text{ReLER Lab, University of Technology Sydney}$
}

\maketitle

\begin{abstract}
   Obtaining viewer responses from videos can be useful for creators and streaming platforms to analyze the video performance and improve the future user experience. In this report, we present our method for 2021 Evoked Expression from Videos Challenge. In particular, our model utilizes both audio and image modalities as inputs to predict emotion changes of viewers. To model long-range emotion changes, we use a GRU-based model to predict one sparse signal with 1Hz. We observe that the emotion changes are smooth. Therefore, the final dense prediction is obtained via linear interpolating the signal, which is robust to the prediction fluctuation.Albeit simple, the proposed method has achieved pearson's correlation score of $0.04430$ on the final private test set.
\end{abstract}

\section{Introduction}
Videos, as rich-media content, are capable of evoking a range of emotions of viewers~\cite{sun2021eev}. With the booming of streaming platforms, creating attractive videos 
has become a key demand for both platforms and creators.  Therefore, the 2021 Evoked Expression from Videos Challenge  intends to solve this challenge and proposes the objective: predicting continuous viewer responses from youtube videos at the rate of 6Hz. Predicting evoked affect before viewers see the video is helpful to future video creation as well as recommendation optimization. 
The organizers provide Internet videos of more than 20 themes covering music, films, games, and more. The average length of these videos is 4.3 minutes which is approximately 1500 predictions to be made. The labels are logits with corresponding timestamps. The predictions of one video are evaluated using pearson's correlation coefficient for each emotion. The final score is averaged over the all $15$ emotions on the private test set, which contains $1377$ videos. 

The proposed model is shown in Figure~\ref{fig:model-arch}. Given one video, we split it into frames and audio inputs. We design one model architecture that exploit the temporal relation within each modality and the cross-modality information by fusing the feature together to generate viewer emotion predictions. As the competition demands, we are required to generate one dense predictions of viewer responses at $6$Hz. Every second we need to predict 6-time emotion changes and the total number of prediction for each video is about $1500$. 
We observe that the emotion changes are smooth. In the experiment, we verify this point that directly predicting the $6$Hz dense prediction is not stable and the pearson's correlation is much lower than the smooth prediction from sparse signal. In particular, the final model is to predict reactions at 1Hz, which reduces the total of each video to about 250 predictions. In the end, we use interpolation techniques to compute the final dense 6Hz predictions.

In the following sections, we will explain our method and several attempts on this task in detail. In Section~\ref{sec:vidrep}, we first study how to encode one video, followed by the proposed reaction prediction model in Section~\ref{sec:recation}. The experimental results are provided in Section~\ref{sec:exps} . The summary and future works are discussed in Section~\ref{sec:conclusion}. 

\section{Video Representation}\label{sec:vidrep}

\subsection{Visual Features}
We deploy the off-the-self Swin Transformer~\cite{liu2021swin}, \ie, Swin-L, to extract image features, which is pre-trained on ImageNet-22K~\cite{krizhevsky2012imagenet}. Each sampled frame is resized to $224\times 224$. We extract the output at stage 4 as the visual feature, which is a feature vector of $1536$ dimensions. In practice, we also can use the feature extracted by SE-ResNet~\cite{hu2018squeeze} and InceptionNet~\cite{szegedy2015rethinking}. The only requirement is to change the output feature dimension for the subsequent training.

\subsection{Audio Features}

The audio features are extracted using VGGish~\cite{vggish}, which is pretrained on YouTube-100M~\cite{vggish}. The audio track of each video is first transcoded into 16kHz mono audio and then using the method from AudioSet~\cite{gemmeke2017audio} to compute the log mel-spectrogram. The $94\times 64$ log mel-spectrogram is then fed into VGGish resulting in a audio representation of 128 dimensions.

\section{Reaction Prediction Model} \label{sec:recation}

\subsection{Model Structure}
\begin{figure}[t]
   \begin{center}
      \includegraphics[width=1.0\linewidth]{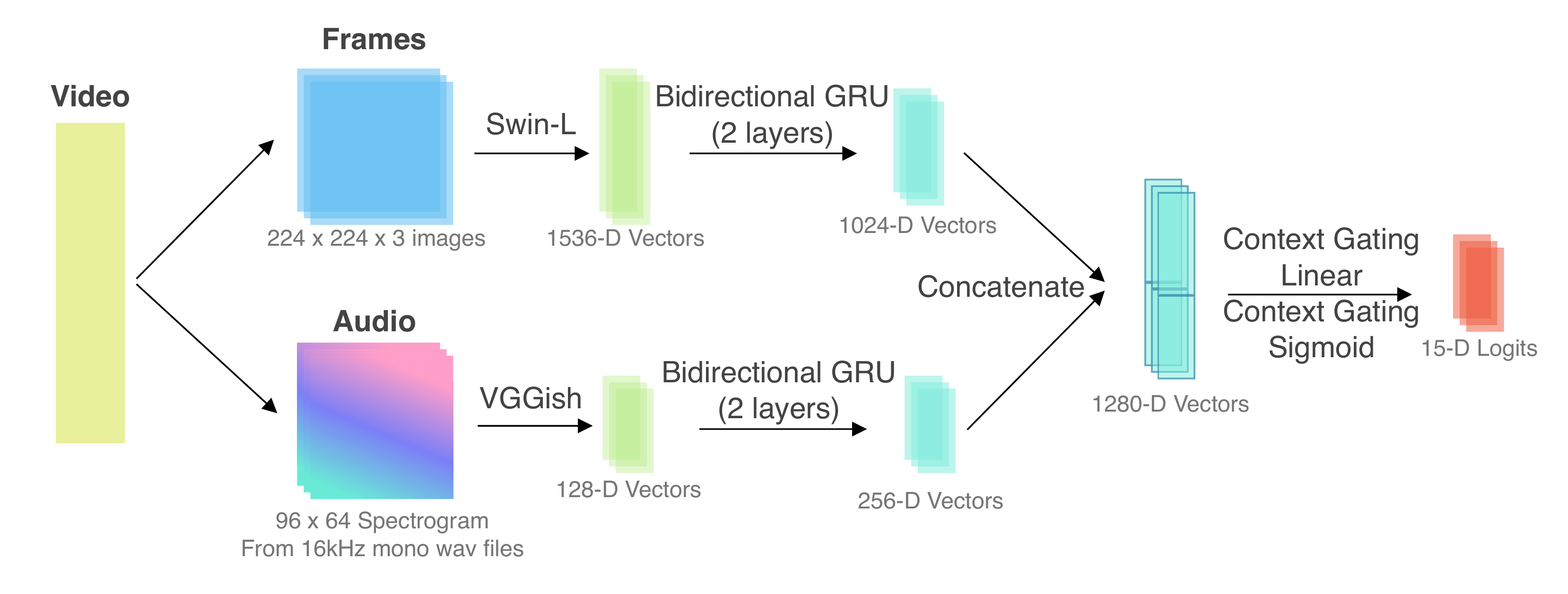}
   \end{center}
   \caption{Our model architecture. Given one video, we first leverage the off-the-shelf networks~\cite{vggish, liu2021swin} to extract the visual and audio features, respectively. Then we apply the bidirectional Gated recurrent unit (GRU) to encode the temporal contextual information. The late fusion strategy is adopted in the proposed framework, and we concatenate the two modality feature and adopt the context gating function to acquire the emotion score of $15$ pre-defined categories.  }
   \label{fig:model-arch}
\end{figure}

Our model is based on the baseline model proposed by Sun \etal~\cite{sun2021eev}, as shown in Figure~\ref{fig:model-arch}. The pre-computed features of each modality are fed into 2-layer bidirectional gated recurrent units (GRUs)~\cite{cho2014learning} without sharing weights. The bidirectional GRU builds the correlation in the temporal space. Comparing to unidirectional GRU, it takes both the past and future information into account, which enables us to form a better generalization at each time step for each modality. It is worth noting that we do not share weight for the two modalities, and the temporal information is only shared within each modality. We adopt the late fusion strategy~\cite{karpathy2014large}. The output of GRUs is concatenated as the video representation. 
Context gating~\cite{miech2017learnable} is used to exploit the dependencies within the fused feature vector. 
In the end, another context gating layer is added along with a sigmoid activation layer to calculate the final prediction. The final emotion scores of every frame are within $[0,1]$, and we notice that the sum of $15$ emotion categories is not supposed to be $1$.

\textbf{Optimization Objective.} 
We use the element-wise $L_1$ loss. 
The $L_1$ loss enforces the model to match the labels at a frame level as well as follow the trend of each expression along the temporal dimension. 

\subsection{Sparse Sampling}

The task requires us to generate dense 6Hz predictions for each video, but we notice the sparse emotion label is more stable. In practise, we train a model based on the 1Hz sparse sampling of each video. Each video is divided into 60 seconds clips and sampled at 1Hz for feature extraction. This is a trade-off between a moderate temporal perception field and the training difficulty of the GRUs. As our GRU module runs for 60 time steps at each run, sampling at 1Hz will provide a perception field of $60$ seconds. For the result submission, we use linear interpolation to generate the final dense 6Hz expression predictions as required.
More discussion on the sparse sampling can be found in Section~\ref{subsec:interpolate}.

\section{Experiments}\label{sec:exps}

\subsection{Dataset}

We use the partial EEV dataset~\cite{sun2021eev} provided by the organizer to train and evaluate our model. The partial dataset consists of 3061 training videos, 755 validation videos, and 1377 test videos. During the challenge data preparation, we notice that some videos are missing due to video unavailability or being private. Therefore, we actually obtain 3023 videos for training and 745 videos for validation. There are more than 20 themes for the videos in the dataset with an average video length of 4.3 minutes. There are in total 15 annotated expressions in the dataset. Several expressions might denote similar basic emotion and differ in degree, like elation and amusement. This also increases the difficulty in predicting the right expression since they are hard to differentiate in nature.

\subsection{Linear Interpolation of Predictions}\label{subsec:interpolate}

We train our model on 60-frame video clip with 1Hz sample rate, which covers a 60-second time span. We consider three different strategies (see Table~\ref{tbl:interpolate}) to obtain the dense prediction at 6Hz for submission. First, we consider to keep the input frame number, \ie, 60 frames, and change the  sampling rate from $1$Hz to $6$Hz. The time span is also reduced from $60$ seconds to $10$ seconds, which compromises the inference result. 
Second, we keep the perception range, \ie, $60$s and sample the frame at $6$Hz, resulting the input of $360$ frames. In experiment, we also find that the dense inputs harm the performance.
Third, we can use the linear interpolation strategy to up-sample the 1Hz predictions to 6Hz. The test setting is close to the training process, and achieves the competitive performance.
For the final submission, we adopt the third strategy and find that the sparse input sampling and the dense prediction interpolation are helpful. 

\begin{table}[]
   \begin{center}
   \begin{tabular}{c|c|c|c}
   \shline
\multirow{2}{*}{Sample rate} & \multicolumn{2}{c|}{Clip} & \multirow{2}{*}{Best Val corr.}\\ \cline{2-3}
                             & Length     &  \#Frames       & \\ \hline
6Hz (wo interp) &   10s      &   60    &     0.0077    \\
6Hz (wo interp) &   60s      &  360    &     0.0106    \\ \hline
1Hz (w interp)  &   60s & 60 &  0.0121 \\
   
   \shline
   \end{tabular}
   \end{center}
   \caption{Ablation Study. We tested different strategies to obtain 6Hz dense predictions as  described in Section~\ref{subsec:interpolate}.}
   \label{tbl:interpolate}
   \end{table}


\subsection{Low-pass Label Filtering}

Here we report one failed attempt. 
From our observation of the expression labels, we find that it contains high-frequency noises and sudden changes to 0 (caused by technical reason when collecting the dataset, as explained in~\cite{sun2021eev}). These factors make the model hard to fit the data. Therefore, one straight-forward idea is to use low-pass filters to filter out these high-frequency noises and ease out the sudden ramps. 
To verify this point, we calculate pearson's correlation coefficient between the filtered label and the original label (see Table~\ref{tbl:filter}). 
If we train the model to fit the filtered label, this label correlation score can be considered as an upper bound of the model trained on the filtered data. 
While the smoothed labels are easier to fit, we observe that the model learned on filtered labels do not achieved better performance than the model learned on the original label (see Table~\ref{tbl:filter}). 

\begin{table}[]
   \begin{center}
   \begin{tabular}{l|c|c}
   \shline
   Filters             & Label corr.   & Trained Model corr.  \\ \hline
   Butterworth Filter  & 0.63          & 0.0089   \\ 
   Median Filter       & 0.66          & 0.0090   \\ 
   Gaussian Filter     & 0.69          & 0.0071   \\ \hline
   Original Labels     & 1.00          & 0.0121   \\ 
   \shline
   \end{tabular}
   \end{center}
   \caption{One failed attempt. We enforce the model to learn the filtered labels, which is smoother and robust to noisy annotations. However, we find the predicted label on training set only obtains worse corr. than the model learned on the original label without smoothing. 
   The correlation between the original label and the filtered training label is listed in the second column of this table. 
   The third column contains the best validation correlation we got using the filtered labels as our training target, and all models use the same hyper-parameters.}
   \label{tbl:filter}
   \end{table}

\subsection{Different Loss Functions}
In our model, we use a simple but yet effective $L_1$ loss. One optional method is to use the KL Divergence Loss to minimize the distance between the predicted logit and the ground-truth label. However, we notice that the expression labels cannot be strictly considered as a probability distribution, since the sum score of $15$ emotions in every frame does not equal to $1$. However, later testing (after the challenge) reveals some interesting results as shown in Table~\ref{tbl:lossexp}. It out performs the $L_1$ loss we use in our final submission on the validation set.

We also have tried another alternative correlation-based loss called concordance correlation coefficient (CCC) (see Equation~\ref{eqn:ccc}). The existing work~\cite{atmaja2021evaluation} has shown that CCC performs better than error-based losses in terms of average CCC score.
\begin{equation}
    \label{eqn:ccc}
    \rho_{c}=\frac{2 \rho \sigma_{x} \sigma_{y}}{\sigma_{x}^{2}+\sigma_{y}^{2}+\left(\mu_{x}-\mu_{y}\right)^{2}}
\end{equation}
$\mu _{x}$ and $\mu _{y}$ are the means for the two variables and $\sigma _{x}^{2}$ and $\sigma _{y}^{2}$ are the corresponding variances. $\rho$  is pearson's correlation coefficient between the two variables. This eliminates the square root part in perason's correlation and makes it easier to optimize. \textbf{In our practise, although CCC is more stable than the $L_1$ loss on the training set, it leads to a worse correlation score on the validation set (see Table~\ref{tbl:lossexp}).}

\begin{table}[]
   \begin{center}
   \begin{tabular}{l|c}
   \shline
   Losses    & Best Validation corr.  \\ \hline
   L1 loss      & 0.0121   \\ 
   \textbf{KL loss}     & \textbf{0.0137}   \\
   CCC loss     & 0.0117   \\
   \shline
   \end{tabular}
   \end{center}
   \caption{Ablation Study. We report the best validation results on L1 Loss, KL loss and CCC loss. KL loss performs best in this test. But since this test is conducted after the challenge, we still use the L1 loss in our final submission.}
   \label{tbl:lossexp}
   \end{table}

\subsection{Final Submission}

The final submitted result is obtained by ensembling $8$ top models on the validation set. Among the eight models, we also include one model trained on both training and validation sets. We achieved a pearson's correlation score of $0.04430$ on the final private test set of the EEV challenge (see Table~\ref{tbl:leaderboard}).

\begin{table}[]
   \begin{center}
   \begin{tabular}{l|c}
   \shline
   Teams    & Final Test corr.  \\ \hline
   SML      & 0.05477   \\ 
   \textbf{Ours}    & \textbf{0.04430}	   \\ 
   youlin   & 0.02292   \\ 
   VAOH     & 0.01510	   \\ 
   jskim    & 0.01402 \\
   \shline
   \end{tabular}
   \end{center}
   \caption{The final correlation score on the private test set of the top 5 teams.}
   \label{tbl:leaderboard}
   \end{table}

\section{Conclusion}\label{sec:conclusion}
In this report, we present our approach for the viewer reaction prediction. Our model takes the advantages of both image and audio modalities to build the temporal correlation. During inference, we use the linear interpolation to generate dense predictions from sparse predictions. Albiet simple, the proposed method have achieved the 2nd place on the EEV Challenge leaderboard. 
In the experiment, we not only illustrate our detailed solution but provide our failed study on different loss terms and label smoothing strategies. We hope it can pave the way for future works on reducing the noise in the labels or a new loss function to regularize the model training. In the future, we will continue to study more discriminative cross-modality losses, such as Instance loss~\cite{zheng2020dual} and Clip loss~\cite{radford2021learning}, and extend the proposed method to more real-world video understanding tasks, such as sign language recognition~\cite{li2020transferring}.

{\footnotesize
\bibliographystyle{ieee_fullname}
\bibliography{egbib}
}

\end{document}